\documentclass{MechatronikTagung}
\usepackage{printlen}
\usepackage{svg}
\usepackage[utf8]{inputenc} 
\usepackage[T1]{fontenc}

\usepackage{amsmath} 
\usepackage{amssymb} 

\usepackage{graphicx}

\usepackage{booktabs} 
\usepackage{cite} 

\titelenglisch{A Hybrid Autoencoder for Robust Heightmap Generation from Fused Lidar and Depth Data for Humanoid Robot Locomotion}

\autorenliste{%
Dennis Bank$^1$,
Joost Cordes$^1$,
Thomas Seel$^1$,
and Simon F.G. Ehlers$^1$, 
\vspace{2mm}\\
$^1$ Institute of Mechatronic Systems, Leibniz University Hannover, 30823 Garbsen, Deutschland;\\
 Contact: dennis.bank@imes.uni-hannover.de}

\kurzfassungenglisch{Reliable terrain perception is a critical prerequisite for the deployment of humanoid robots in unstructured, human-centric environments. While traditional systems often rely on manually engineered, single-sensor pipelines, this paper presents a learning-based framework that uses an intermediate, robot-centric heightmap representation. A hybrid Encoder-Decoder Structure (EDS) is introduced, utilizing a Convolutional Neural Network (CNN) for spatial feature extraction fused with a Gated Recurrent Unit (GRU) core for temporal consistency. The architecture integrates multimodal data from an Intel RealSense depth camera, a LIVOX MID-360 LiDAR processed via efficient spherical projection, and an onboard IMU. Quantitative results demonstrate that multimodal fusion improves reconstruction accuracy by 7.2\% over depth-only and 9.9\% over LiDAR-only configurations. Furthermore, the integration of a 3.2~s temporal context reduces mapping drift.}
\begin{document}

\section{Introduction}

The achievement of reliable and safe locomotion in unstructured environments remains a fundamental challenge in the field of humanoid robotics. While bipedal systems offer high versatility for tasks in human-centered environments, locomotion stability is inherently compromised by a high center of mass and bipedal morphology \cite{TLZ24}. To navigate these complexities, control policies are required that effectively integrate proprioception with exteroceptive perception, such as information from depth cameras or Light Detection and Ranging (LiDAR) sensors \cite{RKAB23}.

Existing approaches to terrain-aware humanoid locomotion frequently rely on single-sensor solutions. However, depth camera data is highly sensitive to lighting and surface properties, while LiDAR processing often introduces significant latency in dynamic environments \cite{RKAB23, MWG+22}. Furthermore, traditional mapping techniques such as Simultaneous Localization and Mapping (SLAM), elevation mapping, and voxelization are often computationally demanding, prone to drift, or reliant on manually engineered preprocessing pipelines that require time and effort to make them robust across all corner cases.

To address these limitations, a learning-based perception-to-control framework is presented, which uses a robot-centric heightmap as an intermediate representation \cite{DPG+24}. This modular design allows the perception module to focus exclusively on reconstructing the environment from raw sensor inputs. The primary contribution of this research involves the development of a hybrid encoder-decoder structure (EDS) that utilizes a convolutional neural network (CNN) for spatial feature extraction and a gated recurrent unit (GRU) for temporal modeling. Robustness is further enhanced through the fusion of complementary data from an Intel RealSense depth camera and a LIVOX MID-360 LiDAR, with the latter processed via an efficient spherical projection into structured 2D range images.

The heightmap representation itself, specifically regarding the spacing and resolution of grid points, is optimized for optimal walking policy behaviour. The impact of grid resolution on the stability and convergence of locomotion policies is systematically evaluated, as the density of these points defines the balance between foothold precision and computational efficiency. It is demonstrated that while finer resolutions enhance the detection of discrete terrain features, excessive dimensionality can lead to training instability. Through this dual focus on hybrid multimodal fusion and tailored grid spacing, a validated foundation is established for real-time, terrain-aware humanoid locomotion in complex environments.

\begin{figure}[h]
    \centering
    \includegraphics[width=1\linewidth]{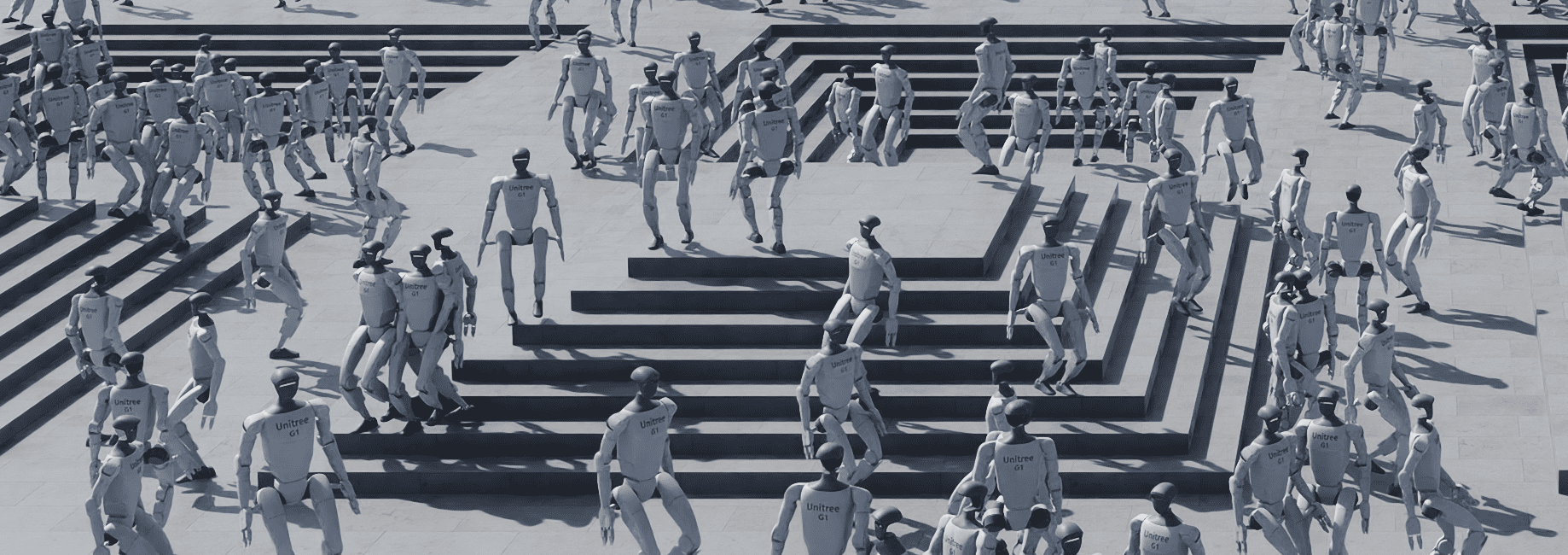}
    \caption{Training environment in Isaac Lab \cite{lab}. Thousands of robots train to walk in different environments in parallel.}
    \label{fig:enter-label}
\end{figure}

\section{Locomotion Policy and Heightmap Optimization}

The locomotion control is realized through a deep reinforcement learning (DRL) policy trained in Isaac Lab \cite{lab} using Proximal Policy Optimization (PPO) \cite{SWD+17} in an actor-critic framework.

To ensure stability and environmental awareness, the policy gets a structured observation space that integrates proprioceptive signals, base velocity commands, and the robot-centric heightmap generated by the perception module as inputs \cite{DPG+24}. The action space consists of target joint positions for the 23 actuated degrees of freedom, which are subsequently tracked by low-level proportional-derivative (PD) controllers \cite{RHRH22}. 

\subsection{Heightmap Representation and Grid Spacing Optimization}

\begin{figure}[h]
    \centering
    \includegraphics[width=\linewidth]{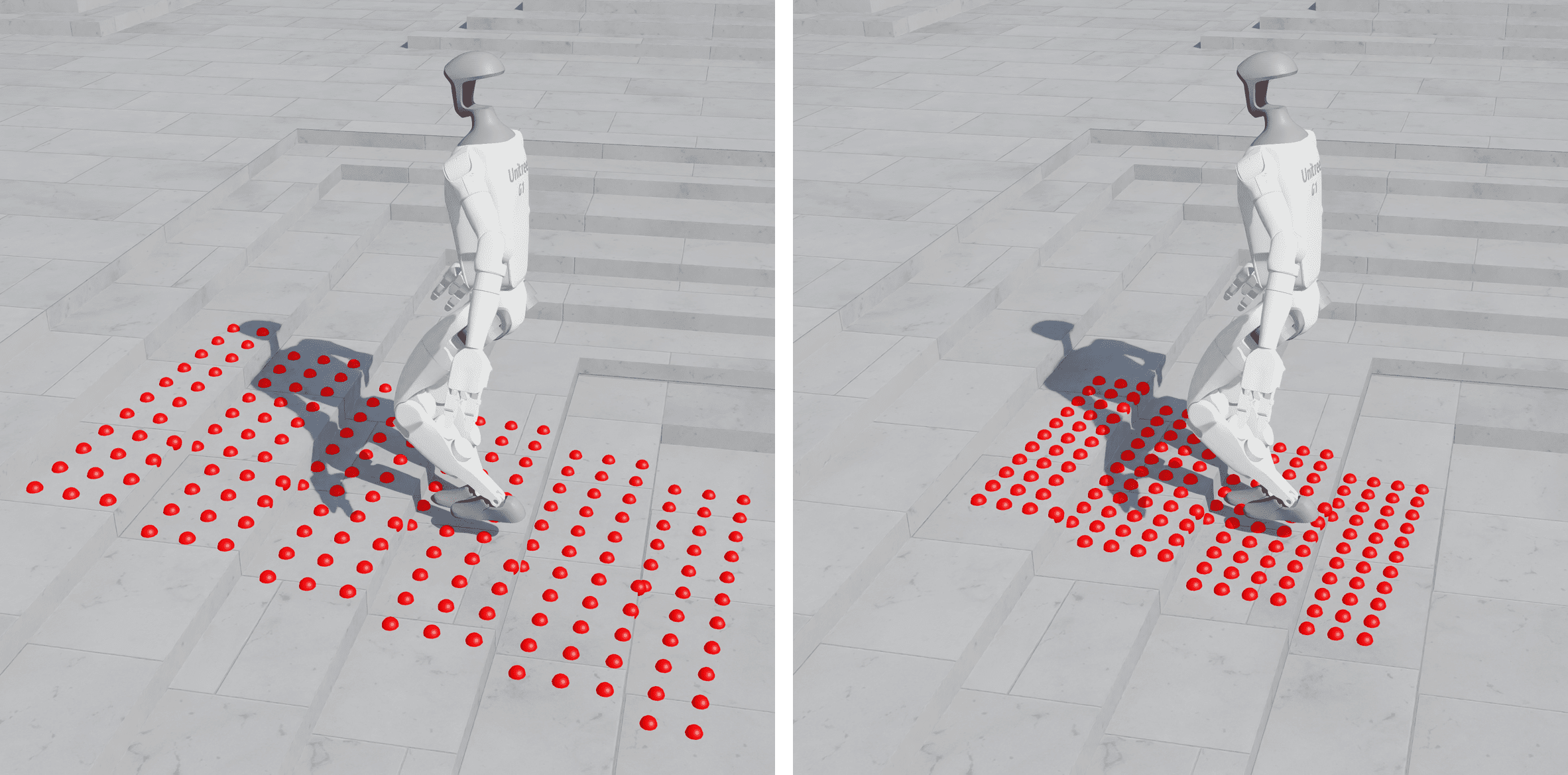}
    \caption{Different heightmap configurations were investigated. The distance between the points, the number of points in the longitudinal and lateral direction, as well as the position of the height map relative to the robot, were optimized.}
    \label{fig:hmopt}
\end{figure}

A critical aspect of the perception-to-control pipeline is the design of the robot-centric heightmap, which serves as the primary exteroceptive input for the policy. The heightmap is generated by projecting terrain data from a virtual scanner aligned above the robot's base link, where each grid cell stores the local elevation relative to the robot base. A comparison of two different heightmap configurations is shown in Figure \ref{fig:hmopt}.

The optimization of grid point spacing and resolution was important for stable locomotion. It was observed that the resolution defines a fundamental trade-off between the precision of terrain feature detection and the computational efficiency of the training process. Systematic investigation identified an optimal resolution range of 6--8~cm per cell. Specifically, a heightmap size of 0.98~m in length and 0.7~m in width, featuring a resolution of 7~cm, demonstrated the most favorable results. To enhance the informativeness of the representation in the robot's immediate path, the heightmap center is shifted forward by 0.2~m, thereby enlarging the field of view in the direction of travel.

Deviations from this spacing were found to negatively impact policy performance. Higher resolutions (finer spacing) increased the dimensionality of the perceptual input, which hindered policy learning and led to training instability. Conversely, lower resolutions (coarser spacing) failed to adequately capture discrete terrain features such as stair edges, leading to reduced foothold precision and a higher frequency of stumbles and falls. These findings demonstrate that a tailored, compact heightmap representation is a prerequisite for the emergence of anticipatory behavior and stable locomotion on complex terrains.

\subsection{Reward Design for Anticipatory Behavior}

To promote terrain-aware behavior within this optimized heightmap framework, the reward function was iteratively refined. Key adjustments included the introduction of a symmetry term for foot airtime and an increased penalty for sliding foot contacts. 

These modifications encouraged the policy to lift feet more deliberately and plan foothold placement based on the upcoming terrain elevation provided by the heightmap. It was established that the combination of optimal grid spacing and refined reward shaping enables the robot to anticipate elevation changes and execute precise, terrain-adapted foot trajectories.

\section{Encoder-Decoder Structure for Heightmap Reconstruction}

\begin{figure}
    \centering
    \includegraphics[width=.6\linewidth]{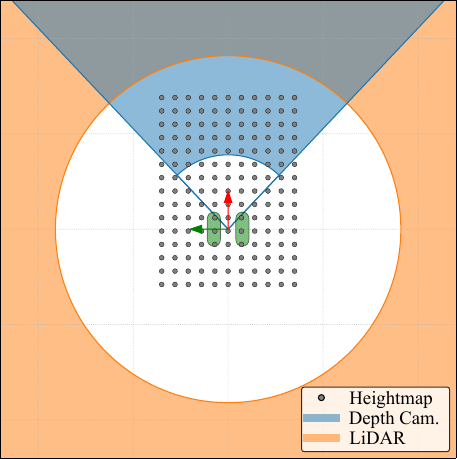}
    \caption{Overview of the heightmap and the different sensor modalities. Substantial parts of the heightmap are not directly covered by the depth camera or the LiDAR and need to be reconstructed using information from the past, when they were in view.}
    \label{fig:fov}
    \vspace{-0.5cm}
\end{figure}

The perception module is required to transform raw, high-dimensional sensory data into the optimized robot-centric heightmap established in the previous section. An encoder-decoder structure is necessary for this task because raw exteroceptive sensor data, i.e., point clouds and depth images, do not naturally align with the intermediate heightmap representation required by the locomotion policy, as shown in Figure \ref{fig:fov}. The structure is designed to bridge this gap by extracting spatial features and integrating them across a temporal horizon to ensure consistent terrain perception \cite{DPG+24, ZŁ21}.

\subsection{Spherical Projection for LiDAR Processing}

To enable efficient processing of unstructured 3D LiDAR point cloud data within a convolutional neural network (CNN) framework, a spherical projection is applied \cite{GZKX24}. This transformation maps Cartesian coordinates $(x, y, z)$ into a range-azimuth-inclination space, which is subsequently discretized into a structured 2D image grid. The resulting representation, referred to as a range image, encodes distance values for each LiDAR beam and allows the network to exploit local spatial correlations similar to standard image data \cite{GZKX24}.

In this framework, the LIVOX MID-360 point cloud is projected into a $276\times40$ resolution grid, where the width is defined by the horizontal angular resolution and the height corresponds to the 40 vertical channels of the sensor. 

The effective vertical field of view (FOV) utilized ranges from $-7^{\circ}$ to $52^{\circ}$, with a horizontal resolution of approximately $1.3^{\circ}$ at an update rate of 10~Hz. To ensure the most efficient data ingestion for the convolutional layers, the seam of the spherical unfolding is positioned at the rear of the robot base.

The projection process inherently introduces nonuniform spatial resolution, where points near the sensor are overrepresented while distant surfaces are compressed. To mitigate artifacts inherent in the projection process, such as ringing or broken ring aliasing, specific preprocessing operations are required. These include row-wise gap-filling and nearest-neighbor filling, each followed by the application of a $3\times3$ median filter \cite{GZKX24}. Furthermore, values are clipped to a practical operating range between 0.2~m and 3.0~m to match the sensor's effective distance for foothold planning. This approach significantly reduces the computational complexity, memory requirements, and processing latency compared to voxel-based or point-based methods, making the perception pipeline scalable for real-time humanoid applications on resource-constrained hardware.

\subsection{Architectural Components and Layer Configuration}

The architecture follows a two-stage design, utilizing separate convolutional neural network (CNN) encoders for each visual modality and a shared recurrent decoder core \cite{DPG+24}. The configuration for each component is detailed as follows:

\begin{itemize}
    \item \textbf{CNN Encoders:} Separate CNN branches process depth images (at $160\times120$ resolution) and spherical LiDAR range images (at $276\times40$ resolution). Each encoder consists of four strided convolutional layers using $3\times3$ kernels, which progressively downsample the spatial resolution while increasing feature depth. The extracted feature maps are flattened and mapped through fully connected layers to generate a 256-dimensional latent representation for each modality.
    
    \item \textbf{Multimodal Fusion:} The 256-dimensional latent vectors from the camera and LiDAR are concatenated with additional auxiliary inputs: a 15-dimensional robot state vector (containing IMU-derived linear and angular velocities, position, and orientation) and the 165-dimensional heightmap prediction from the previous timestep. These heterogeneous inputs are fused into a unified multimodal representation via linear transformations and layer normalization.
    
    \item \textbf{Recurrent Decoder:} The temporal core comprises two stacked Gated Recurrent Unit (GRU) layers, each with a hidden size of 256. The GRUs capture temporal dependencies and short-term motion dynamics, which stabilize the terrain predictions when sensors are occluded or data is noisy \cite{ZŁ21}. The resulting hidden state is passed to a decoder head consisting of two fully connected layers that output the final 165-dimensional predicted heightmap \cite{DPG+24}.
\end{itemize}

\begin{figure}[h]
    \centering
    \includegraphics[width=1\linewidth]{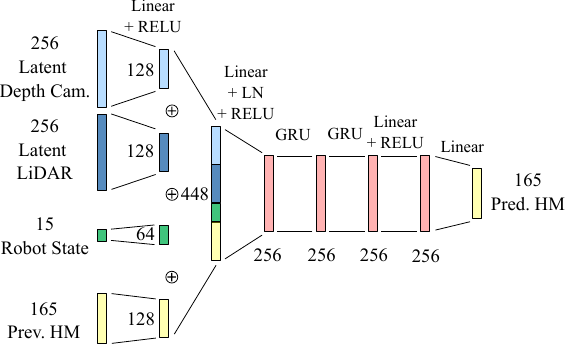}
    \caption{EDS used to predict the heightmap. It consists of pretrained encoders that compress the information from the depth camera and LiDAR. Furthermore, it takes the current robot state as well as the previous heightmap to predict the current heightmap.}
    \label{fig:eds}
\end{figure}

An overview of the EDS structure is shown in Figure \ref{fig:eds}.

\subsection{Two-Stage Training Process}

A staged training methodology is employed to ensure the stability of the feature extraction process and the accuracy of the final terrain reconstruction. This separation of learning phases allows for more targeted improvements and better interpretability compared to monolithic end-to-end pipelines \cite{DPG+24}.

\begin{figure}[h]
    \centering
    \includegraphics[width=1\linewidth]{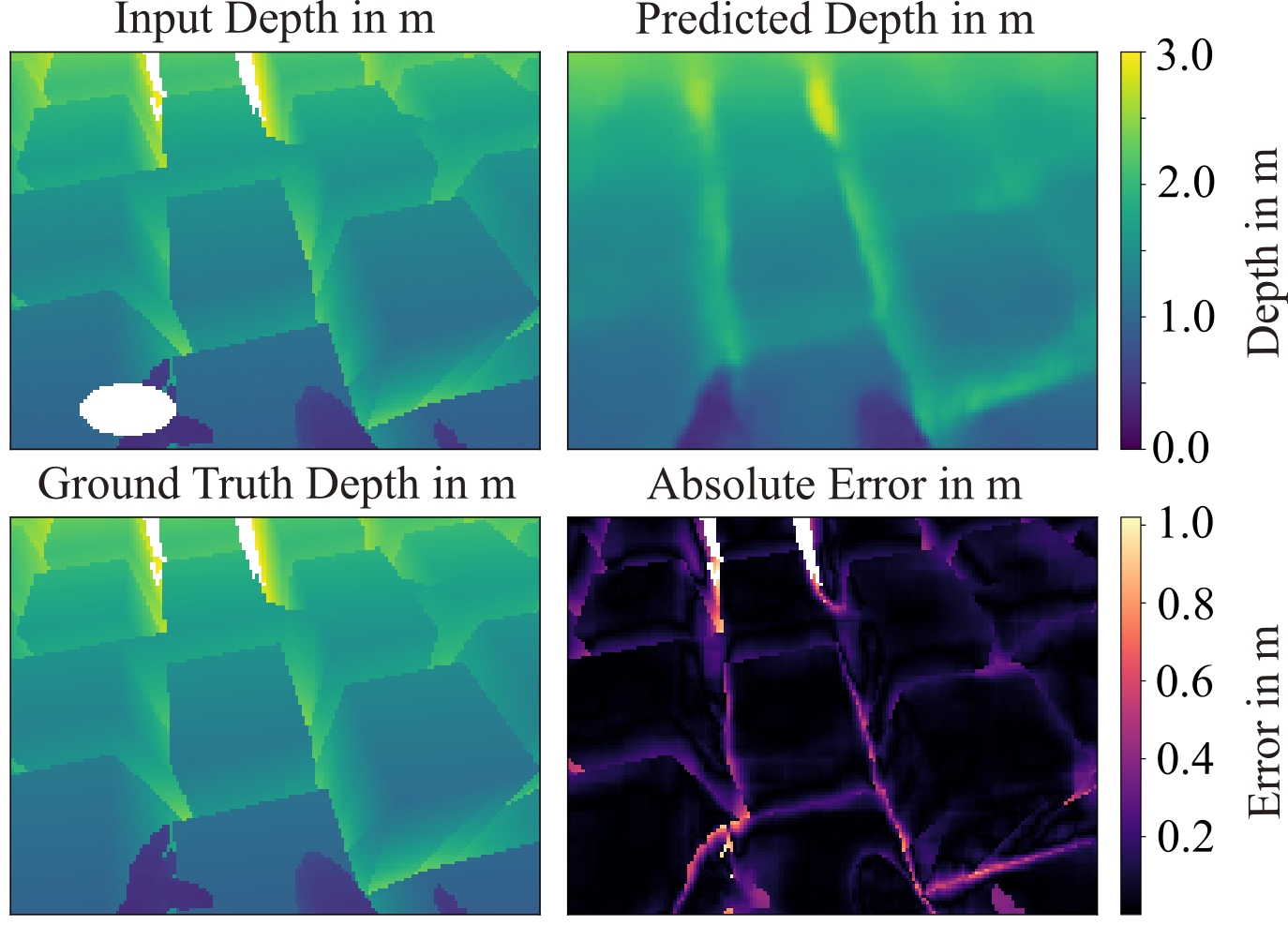}
    \caption{Evaluation of the pretrained encoder for the depth camera. The original image can be reconstructed well, showing only small errors at the edges of objects.}
    \label{fig:error}
\end{figure}

In the first stage, symmetric convolutional autoencoders (AE) are pretrained in an unsupervised manner using noisy simulated sensor data from depth camera and LiDAR modalities. The central purpose of this pretraining is to stabilize feature extraction and ensure that terrain-relevant information is encoded into a compact 256-dimensional latent bottleneck before the introduction of more complex temporal fusion \cite{DPG+24}. To ensure robustness against real-world artifacts, input data is corrupted with Gaussian noise featuring a standard deviation of 1~cm and random-shaped occlusion masks covering up to 3\% of the input image area. Training is conducted using pixel-wise mean squared error (MSE) loss, which was found to consistently outperform Mean Absolute Error (MAE), Huber, and BerHu loss in preliminary reconstruction trials. Notably, U-Net style skip connections are intentionally excluded from this design to force all essential terrain features through the latent bottleneck, thereby enforcing high information density within the representation \cite{DPG+24}. An example of the pretrained AE of the depth camera is shown in Figure \ref{fig:error}.

In the second stage, the pretrained CNN encoders are integrated into the final encoder-decoder structure (EDS), while the original AE decoders are replaced by a recurrent temporal core. This core consists of two stacked Gated Recurrent Unit (GRU) layers with a hidden size of 256, chosen for their favorable balance between representation capacity and inference speed on embedded hardware. The perception module is trained in a supervised setting  with backpropagation through time using ground-truth heightmaps from simulation as the target. 

This process utilizes a comprehensive dataset of 400,000 samples per modality, split into a 70/15/15 ratio for training, validation, and testing. The optimization utilizes the AdamW optimizer with a plateau-based learning rate schedule to achieve stable convergence over 40 training epochs. The resulting framework is designed to operate at 10~Hz, facilitating reliable and high-frequency terrain updates for the locomotion policy.

\section{Results and Discussion}

The evaluation of the proposed framework focuses on the reconstruction accuracy of the perception module and the resulting stability of the locomotion policy. The experiments aim to quantify the benefits of multimodal sensor fusion and the effectiveness of the optimized heightmap grid spacing.

\subsection{Perception Module Performance}

\begin{figure}[h]
    \centering
    \includegraphics[width=1\linewidth]{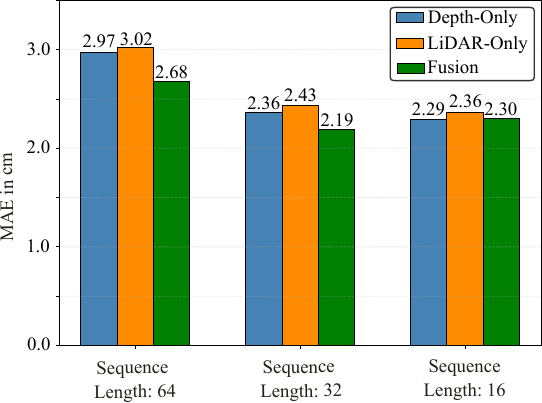}
    \caption{Comparison of MAE of EDS with different inputs. The fusion of LiDAR and depth data provides more accurate heightmap reconstructions than the EDS using only one of the sensor inputs.}
    \label{fig:overviewerror}
\end{figure}

\begin{figure}[h]
    \centering
    \includegraphics[width=1\linewidth]{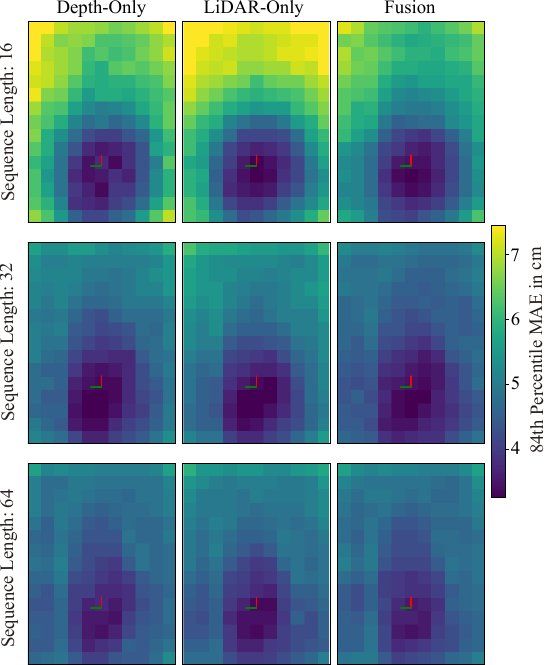}
    \caption{Spatial error distribution around the robot. Notably, the error drops close to zero near the origin, as the robot uses proprioceptive information from its joint angles to determine the ground height under its feet.}
    \label{fig:spatial_error}
\end{figure}

\begin{figure*}[h]
    \centering
    \includegraphics[width=0.9\linewidth]{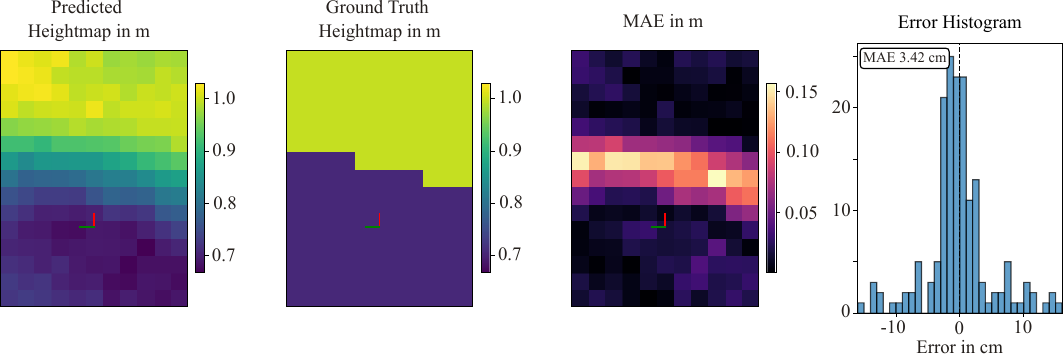}
    \caption{Comparison of a predicted heightmap and the groundtruth close to a step. The elevation change is well reconstructed. However, the edge is smoothed out a little bit, demonstrating a low-pass behavior of the EDS.}
    \label{fig:smooth}
\end{figure*}

The reconstruction accuracy of the Encoder-Decoder Structure was evaluated using the Mean Absolute Error across different sensor modalities and temporal sequence lengths. The results are shown in Figures \ref{fig:overviewerror} and \ref{fig:spatial_error} and can be summarized to:  

\begin{itemize}
    \item \textbf{Sensor Fusion Benefits:} The multimodal fusion setup achieved a reconstruction accuracy of 2.19~cm MAE. This represents a 7.2\% improvement over the depth-only configuration (2.36~cm) and a 9.9\% improvement over the LiDAR-only configuration (2.43~cm). These results confirm that the integration of complementary sensors enhances terrain perception robustness.
    \item \textbf{Temporal Context:} The influence of sequence length proved substantial. Models trained with a moderate temporal horizon of 3.2~s (32 timesteps) achieved approximately 30\% lower reconstruction errors compared to those using short horizons. 
    
    However, increasing the sequence length to 6.4~s yielded diminishing returns, suggesting that a 3.2~s window provides an optimal balance between accuracy and computational efficiency. 
    \item \textbf{Terrain-Specific Accuracy:} High reconstruction accuracy (MAE < 2~cm) was consistently maintained on flat or gently varying surfaces. However, accuracy decreased on discontinuous surfaces such as stairs, where sharp height changes were smoothed into gradual slopes due to the pixel-wise loss function. An example of this smoothing is given in Figure \ref{fig:smooth}.
\end{itemize}

\subsection{Policy Performance and Heightmap Impact}

The locomotion policy was evaluated to determine if a compact, robot-centric heightmap provided sufficient context for terrain-aware control.

\begin{figure}[h]
    \centering
    \includegraphics[width=0.9\linewidth]{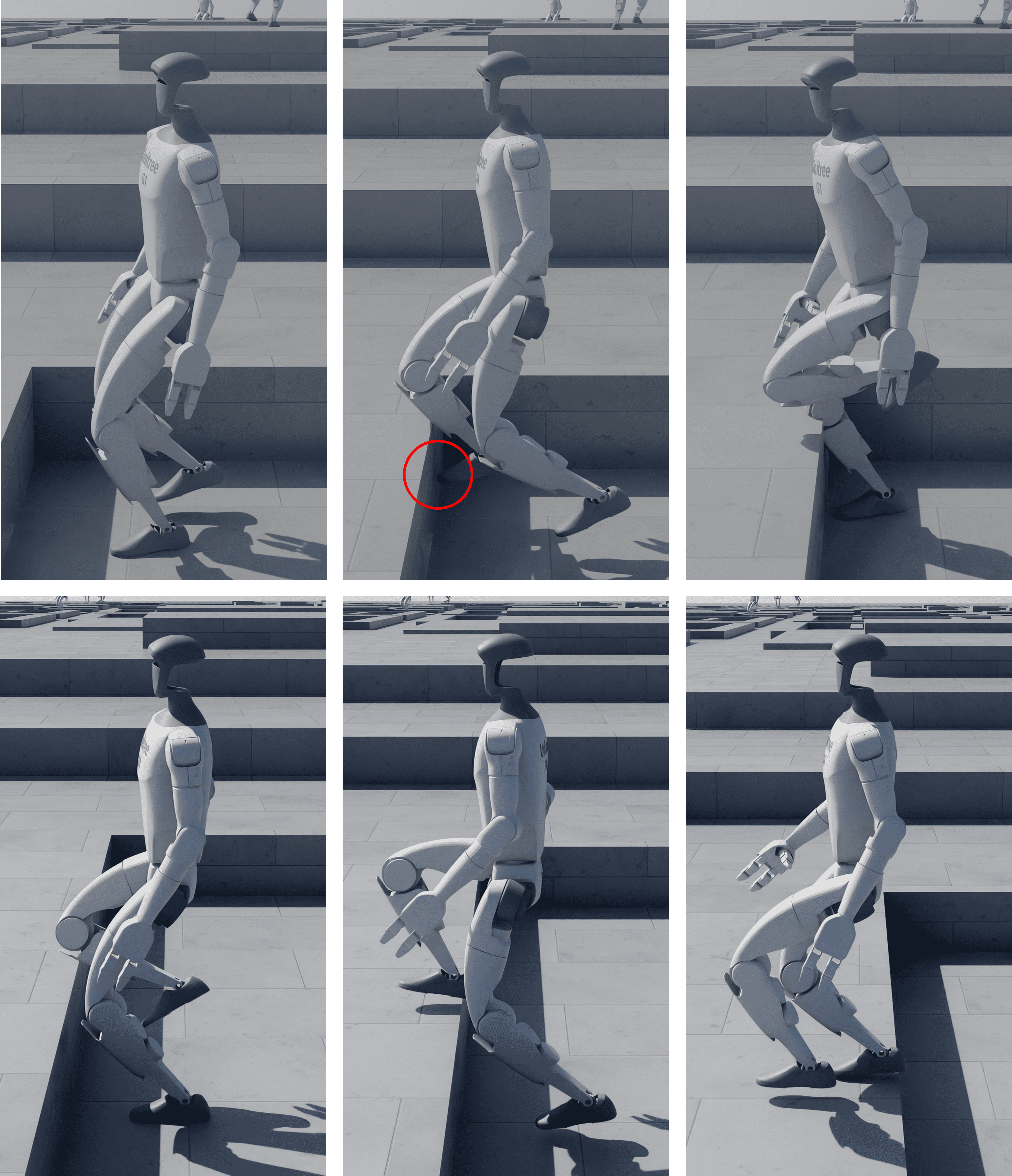}
    \caption{Locomotion without a heightmap (top) vs using a heightmap (bottom). The robot pulls up its leg in anticipation of a step, preventing a collision, as it is able to perceive the ground around it.}
    \label{fig:anti}
    \vspace{-0.5cm}
\end{figure}

\begin{itemize}
    \item \textbf{Anticipatory Behavior:} The customized heightmap (0.98~m $\times$ 0.7~m at 7~cm resolution) enabled the emergence of anticipatory gait patterns as shown in Figure \ref{fig:anti}. Unlike the baseline setup, the optimized policy learned to lift the swing leg precisely to clear upcoming elevation changes, such as steps and stairs.
    \item \textbf{Stability Metrics:} The transition from the blind baseline setup to the customized heightmap and reward structure resulted in a 70.1\% reduction in terminations caused by falls. Furthermore, command tracking precision improved, with a 25.0\% reduction in linear velocity error and a 17.2\% reduction in angular velocity error.
    \item \textbf{Noise Robustness:} The policy demonstrated resilience to perceptual inaccuracies, maintaining stable locomotion even when noise with a standard deviation of up to 2~cm was applied to the heightmap inputs. 
    
    This validates that the compact heightmap representation inherently supports robust control under moderate uncertainty.

\end{itemize}

\subsection{Discussion of Limitations}
The hybrid EDS accurately reconstructs global topology but smooths sharp vertical discontinuities like stair edges due to the pixel-wise MSE loss favoring global statistical fitting over high-frequency spatial details. This represents abrupt height changes as gradual slopes, which may diminish foothold precision. Locomotion stability is also sensitive to reconstruction fidelity; Gaussian noise substantially exceeding a 2~cm standard deviation on heightmaps serves as a practical limit for reliable control \cite{ZYZ24}. Beyond this, anticipatory behavior degrades significantly, increasing fall rates on discontinuous terrain \cite{ZYZ24}. Additionally, the 2.5-D representation cannot capture overhanging structures, and a sim-to-real gap persists, as evidenced by pretraining failures on cluttered real-world datasets. Addressing these challenges requires edge-preserving losses, comprehensive domain randomization \cite{MSD+24}, and hardware-based closed-loop evaluations to validate the impact of latency and synchronization on the integrated perception-to-control pipeline \cite{DPG+24}.

\section{Conclusion and Outlook}

This research presented a learning-based perception-to-control framework for terrain-aware humanoid locomotion, focusing on the development of a hybrid encoder-decoder structure (EDS) and the optimization of robot-centric heightmap representations. By decoupling perception from control, a modular system was established that transforms high-dimensional, multimodal sensory data into a compact format suitable for real-time robotic applications.

The investigation into heightmap grid spacing identified an optimal resolution of 7~cm per cell within a 0.98~m $\times$ 0.7~m grid, shifted forward by 0.2~m to maximize informativeness in the direction of travel. It was demonstrated that this specific spacing balances the requirement for terrain detail with the need for computational efficiency, enabling the locomotion policy to achieve a 70.1\% reduction in fall-related terminations compared to baseline configurations.

The hybrid EDS perception module further validated the efficacy of multimodal sensor fusion. The integration of depth camera and LiDAR data, combined with a GRU-based temporal core, achieved a reconstruction accuracy of 2.19~cm MAE. This setup yielded an improvement of 7.2\% over depth-only and 9.9\% over LiDAR-only modalities, confirming that sensor fusion enhances robustness against individual sensor weaknesses. Furthermore, a moderate temporal sequence length of 3.2~s was found to provide the best balance between reconstruction accuracy and inference efficiency.

While the framework demonstrated stability on flat and gently varying terrains, limitations persist regarding the preservation of sharp discontinuities, such as stair edges, which are often smoothed by the current pixel-wise loss function. Future work should focus on incorporating edge-aware loss functions or hybrid CNN-Transformer architectures to improve high-frequency feature reconstruction. Additionally, the transition toward full closed-loop training on predicted rather than noisy ground-truth heightmaps remains a critical step for enhancing sim-to-real transfer. Ultimately, the validated foundation established in this work provides a scalable path toward autonomous humanoid operation in complex, unstructured environments.

\section*{Acknowledgments}
This study was funded through the transformation network neu/wagen. 
The initiative neu/wagen is funded by the German Federal Ministry for Economic Affairs and Energy (BMWE).\\ 

\begin{figure}[h]
    \centering
    \vspace{-0.5cm}
    \includegraphics[width=0.25\textwidth]{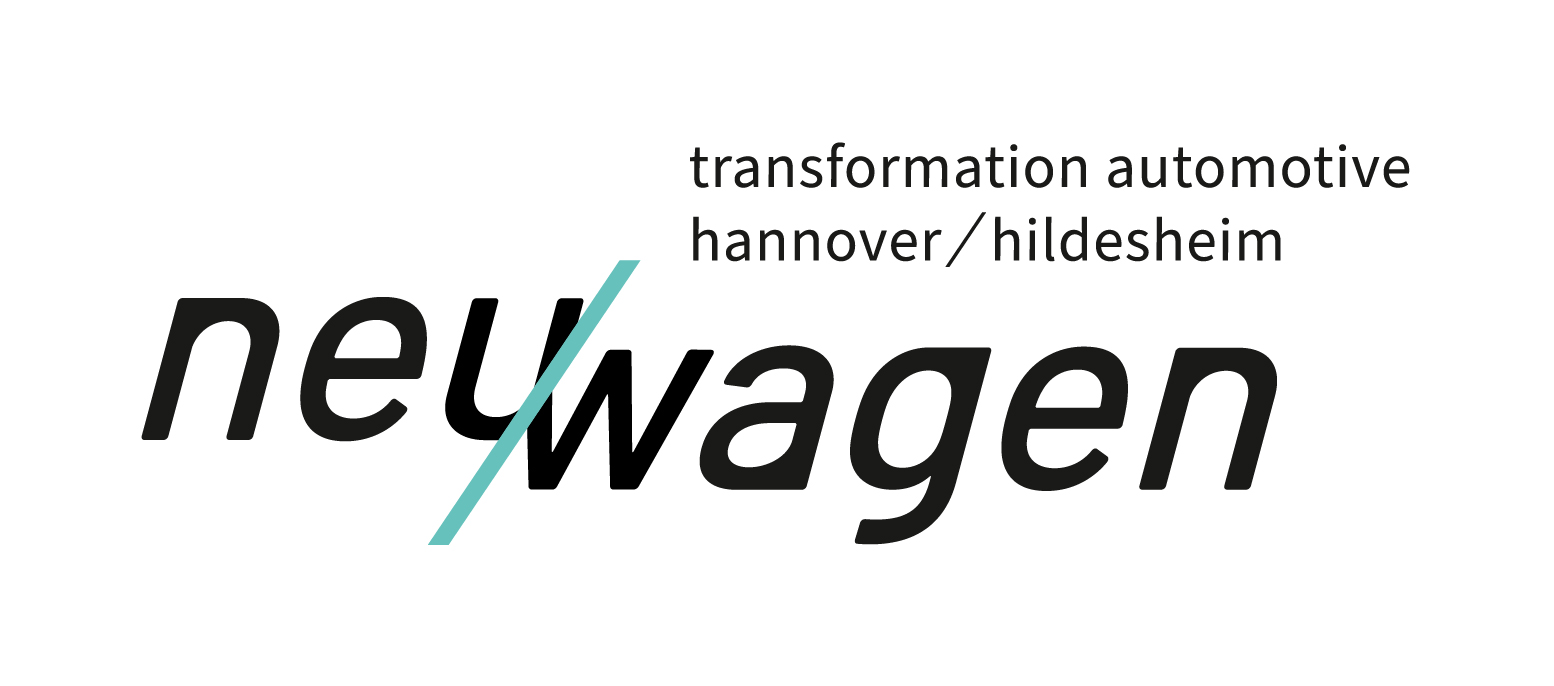}
    \vspace{-0.5cm}
\end{figure}

\begin{figure}[h]
    \centering
    \vspace{-0.5cm}
    \includegraphics[width=0.48\textwidth]{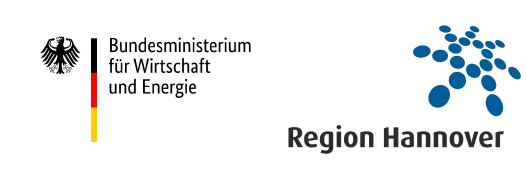}
    \vspace{-1cm}
\end{figure}

\end{document}